\pdfoutput=1

\documentclass[11pt]{article}

\usepackage[]{ACL2023}

\usepackage{times}
\usepackage{latexsym}

\usepackage[T1]{fontenc}

\usepackage[utf8]{inputenc}

\usepackage{microtype}

\usepackage{inconsolata}
\usepackage{comment}

\usepackage{amssymb}
\usepackage{amsmath}
\usepackage{pifont}
\usepackage{color}
\usepackage{xcolor}
\newcommand{\cmark}{\scalebox{1}{\color{blue}\ding{51}}}
\newcommand{\xmark}{\scalebox{1}{\color{magenta}\ding{55}}}
\usepackage[textsize=scriptsize]{todonotes}

\usepackage{CJKutf8}
\newcommand*{\Ja}[1]{%
  \begin{CJK}{UTF8}{ipxm}#1\end{CJK}}

\usepackage{multirow}
\usepackage{booktabs}
\usepackage{hyperref}
\usepackage{breakurl}

\title{Constructing Multilingual Code Search Dataset Using Neural Machine Translation}

\author{Ryo Sekizawa$^1$ \ Nan Duan$^2$ \ Shuai Lu$^2$ \ Hitomi Yanaka$^1$ \\
    $^1$The University of Tokyo\hspace{0.3cm}$^2$Microsoft Research Asia \\
    \texttt{\{ryosekizawa,hyanaka\}@is.s.u-tokyo.ac.jp} \\
    \texttt{\{nanduan,shuailu\}@microsoft.com} \\
}

\begin{document}
\maketitle
\begin{abstract}
Code search is a task to find programming codes that semantically match the given natural language queries. 
Even though some of the existing datasets for this task are multilingual on the programming language side, their query data are only in English. 
In this research, we create a multilingual code search dataset in four natural and four programming languages using a neural machine translation model.
Using our dataset, we pre-train and fine-tune the Transformer-based models and then evaluate them on multiple code search test sets.
Our results show that the model pre-trained with all natural and programming language data has performed best in most cases. 
By applying back-translation data filtering to our dataset, we demonstrate that the translation quality affects the model's performance to a certain extent, but the data size matters more.
\end{abstract}

\section{Introduction}
Code search is the task of finding a semantically corresponding programming language code given a natural language query by calculating their similarity.
With the spread of large-scale code-sharing repositories and the rise of advanced search engines, high-performance code search is an important technology to assist software developers.
Since software developers worldwide search for codes in their native language, we expect code search models to be multilingual.
Although many previous studies focus on multilingual code tasks other than code search (e.g., code generation, code explanation)~\cite{wangCodeT5IdentifierawareUnified2021, ahmadUnifiedPretrainingProgram2021, friedInCoderGenerativeModel2023, zhengCodeGeeXPreTrainedModel2023}, the existing code search datasets ~\cite{husain_codesearchnet_2020,huang_cosqa_2021,lu_codexglue_2021} contain only monolingual data for search queries.

In this research, we construct a new multilingual code search dataset by translating natural language data of the existing large-scale dataset using a neural machine translation model.
We also use our dataset to pre-train and fine-tune the Transformer~\cite{vaswani_attention_2017}-based model and evaluate it on multilingual code search test sets we create.
We show that the model pre-trained with all natural and programming language data performs best under almost all settings.
We also analyze the relationship between the dataset's translation quality and the model's performance by filtering the fine-tuning dataset using back-translation.
Our model and dataset will be publicly available at \url{https://github.com/ynklab/XCodeSearchNet}. 
The contributions of this research are as follows:
\begin{enumerate}
    \setlength{\itemsep}{0.05cm}
    \item Constructing the large code search dataset consisting of multilingual natural language queries and codes using machine translation.
    \item Constructing the multilingual code search model and evaluating it on a code search task using our dataset.
    \item Analyzing the correlation between translation quality and the model performance on a code search task.
\end{enumerate}

\section{Background}
\label{sec:background}
\subsection{Code Search Dataset}
\label{sec:codesearch_dataset}
CodeSearchNet Corpus\footnote{\url{https://github.com/github/CodeSearchNet}}~\citep[CSN;][]{husain_codesearchnet_2020} is a set of code data (\textbf{code}) in six programming languages: Go, Python, Java, PHP, Ruby, and Javascript, and natural language data describing them (\textbf{docstring}).
CSN is created by automatically collecting pairs of function code and its documentation that are publicly available on GitHub and permitted for redistribution.
This corpus contains approximately 2.3 million data pairs and 4 million code-only data.
The natural language data in CSN is function documentation, which is pseudo data of the texts humans use to search for codes.

In contrast, several datasets are created based on natural language queries used for code search by humans.
CodeXGLUE~\cite{lu_codexglue_2021}, a benchmark for various code understanding tasks, includes two code search datasets: WebQueryTest (WQT) and CoSQA~\cite{huang_cosqa_2021}. 
The query data of these datasets are collected from the users' search logs of Microsoft Bing and the code from CSN.
Given these separately collected data, annotators who have programming knowledge manually map the corresponding query and code to construct the dataset.
The common feature of these datasets is that all natural language data, such as docstrings and queries, are limited to English and do not support multiple languages.
\begin{table}[t]
    \small\centering
    \begin{tabular}{lrr} 
        \toprule
                & Pre-training (MLM) & Fine-tuning \\ 
        \midrule 
        PHP     & 662,907 & 1,047,406 \\
        Java    & 500,754 & 908,886 \\
        Python  & 458,219 & 824,342 \\
        Go      & 319,256 & 635,652  \\
        JavaScript & 143,252 & 247,773  \\
        Ruby & 52,905 & 97,580 \\
        \bottomrule
    \end{tabular}
    \caption{Training data size of CSN for each programming language used for pre-training CodeBERT with MLM and fine-tuning on the  code search task.}
    \label{table:dataset_size_codebert}
\end{table}
\subsection{CodeBERT}
\label{sec:codebert}
CodeBERT~\cite{feng_codebert_2020} is a model pre-trained and fine-tuned with CSN and is based on the RoBERTa~\cite{liu_roberta_2019}'s architecture.
CodeBERT uses Masked Language Modeling~\citep[MLM;][]{devlinBERTPretrainingDeep2019b,lampleandconneau_crosslingualpretrain_2019} and Replaced Token Detection~\citep[RTD;][]{clarkELECTRAPretrainingText2020} as pre-training tasks.
Both docstring and code data in CSN are used in MLM, while only code data are used in RTD.
CodeBERT is trained only with English data, thus not available for a code search task with multilingual queries.

\section{Dataset Construction Using Machine Translation}
\label{sec:dataset_construction}
A possible way to construct a code search dataset for multiple languages is to translate an existing monolingual dataset.
However, CSN's large data size makes manually translating all of its docstrings difficult.
Table~\ref{table:dataset_size_codebert} shows the number of CSN data pairs used for pre-training (MLM) and fine-tuning the CodeBERT.

Therefore, we use a machine translation model to translate the English-only data to generate multilingual data efficiently.
By translating CSN docstrings, we create a multilingual dataset consisting of four natural languages (English, French, Japanese, and Chinese) and four programming languages (Go, Python, Java, and PHP).
We also translate the queries in the datasets \citet{feng_codebert_2020} used for fine-tuning and evaluating CodeBERT for our experiments in Section~\ref{sec:training} and Section~\ref{sec:evaluation}.
In their fine-tuning data, the numbers of positive and negative labels are balanced.
Note that we do not use JavaScript and Ruby data, whose sizes are much smaller than those of other programming languages.

As a translation model, we use M2M-100~\cite{fan_beyond_2022}, which supports translations in 100 languages.\footnote{We compared the translation results of some docstrings by several translation models, including Opus-MT and mBART, and chose M2M-100, which achieved the best performance.}
M2M-100 achieved high accuracy in translations of low-resource languages by classifying 100 languages into 14 word families and creating bilingual training data within those families.
We use \texttt{m2m\_100\_1.2B} model, which is provided by EasyNMT\footnote{\url{https://github.com/UKPLab/EasyNMT}}, a public framework of machine translation models.
We set the model's beam size to 3.

We manually annotate the labels to some data of our fine-tuning dataset to check the correlation with the original labels, which is found to be 0.911 (see Appendix~\ref{appendix:dataset_translation} for the details).
\begin{table}[t]
    \small\centering
    \tabcolsep = 0.08cm
    \begin{tabular}{lcccccc} 
        \toprule
                & \multicolumn{3}{c}{Pre-training} & \multicolumn{2}{c}{Fine-tuning} & \multirow{2}{*}{Test} \\ 
                \cmidrule(lr){2-4}\cmidrule(lr){5-6}
                & Train & Valid & Test & Train & Valid \\ 
        \midrule 
        Go      & 316,058&3,198&28,533 & 635,652&28,482& 14,277   \\
        Python  & 453,623&4,596&45,283 & 824,341&46,212 & 22,092  \\
        Java    & 495,768&4,986&42,237 & 908,885&30,654 & 26,646 \\
        PHP     & 656,277&6,630&54,406 & 1,047,403&52,028  & 28,189 \\ \bottomrule
    \end{tabular}
    \caption{The sizes of CSN data for training and evaluating the models in our baseline experiments.}
    \label{table:dataset_size_csn}
\end{table}

\begin{table*}[t]
    \small\center
    \begin{tabular}{clcccccccccccc}
    \toprule
      & & \multicolumn{4}{c}{CSN} & CoSQA & WQT \\ 
    \cmidrule(lr){3-6}\cmidrule(lr){7-7}\cmidrule(lr){8-8}
    & & Go & Python & Java & PHP & Python & Python \\ 
    \midrule
    \multirow{4}{*}{\textbf{No-pre-training}} 
                                & EN & .813 & .801 & .737 & .759 & \textbf{.526} & .334 \\ 
                                & FR & .780 & .708 & .681 & .691 & \textbf{.463} & .302 \\ 
                                & JA & .792 & .686 & .641 & .657 & .372 & .311 \\ 
                                & ZH & .772 & .660 & .633 & .670 & .337 & .297 \\ 
    \midrule
    \multirow{4}{*}{\textbf{All-to-One}} 
                                & EN & .824  & \textbf{.851} & .763 & .790   & .494 & \textbf{.360} \\ 
                                & FR & .798  & \textbf{.796} & \textbf{.733} & .734 & .432 & \textbf{.363} \\ 
                                & JA & .805 & \textbf{.781} & .700 & .711 & \textbf{.460} & .348 \\ 
                                & ZH & .788 & \textbf{.759} & .712 & .731 & \textbf{.427} & \textbf{.359} \\ 
    \midrule
    \multirow{4}{*}{\textbf{All-to-All}} 
                                & EN & \textbf{.835} & .848 & \textbf{.786} & \textbf{.809} & .473 & .351 \\ 
                                & FR & \textbf{.808} & .788 & .731 & \textbf{.759} & .420 & .346 \\ 
                                & JA & \textbf{.816} & .778 & \textbf{.719} & \textbf{.730} & .436 & \textbf{.364} \\ 
                                & ZH & \textbf{.804} & \textbf{.759} & \textbf{.750} & \textbf{.745} & .418 & \textbf{.359} \\
    \bottomrule
    \end{tabular}
    \caption{MRR scores of models pre-trained with all natural language data with either one programming language data or all programming language data.
    }
    \label{table:mrr_baseline}
\end{table*}

\begin{table}[h]
    \small\center
    \begin{tabular}{lcccc}
        \toprule
         & Go &  Python & Java & PHP \\ 
        \midrule
        RoBERTa & .820 & .809& .666& .658 \\
        CODEONLY, INIT=S & .793 &.786 &.657& .617 \\
        CODEONLY, INIT=R & .819 &.844 &.721& .671 \\
        \midrule
        MLM, INIT=S &  .830 & .826 & .714 & .656 \\
        MLM, INIT=R &  .838 & .865 & .748 & .689 \\
        RTD, INIT=R &  .829 & .826 & .715 & .677 \\
        MLM+RTD, INIT=R & .840 & .869 & .748 & .706 \\
        \bottomrule
    \end{tabular}
    \caption{MRR scores of CodeBERT from~\citet{feng_codebert_2020} for Go, Python, Java, and PHP. CODEONLY is RoBERTa pre-trained only with code data. INIT refers to how the parameters of the model are initialized. S is for training from scratch, and R is for the initialization with those of RoBERTa~\cite{liu_roberta_2019}.}
    \label{table:codebert_baseline}
\end{table}

\section{Baseline Experiments}
\label{sec:baseline_experiments}
We conduct baseline experiments, where we train the Transformer-based model with our multilingual dataset under various settings of the data sizes and evaluate it on multiple code search test sets.
\subsection{Training}
\label{sec:training}
We perform pre-training and fine-tuning on a model initialized with the XLM-R~\cite{conneau_xlmroberta_2019} architecture and parameters.
XLM-R is a model pre-trained by MLM with the Wikipedia and Common Crawl corpora for 100 languages using Transformer~\cite{vaswani_attention_2017} and achieved high performance on multilingual tasks, such as question answering.
Note that we use the term ``pre-training'' to refer to further training of XLM-R with our dataset.
In this paper, we use MLM as the learning objective to pre-train XLM-R and then fine-tune it using data pairs whose query and code languages are monolingual. 
We use monolingual data pairs for fine-tuning instead of a multilingual combination, given that~\citet{feng_codebert_2020} clarifies that fine-tuning CodeBERT with six programming languages altogether ``performs worse than fine-tuning a language-specific model for each programming language.''
Query and code data are concatenated to be input to the model, and it predicts their similarity based on the vector representation of the output [\textit{CLS}] tokens.
See Appendix~\ref{appendix:training_settings} for more details on training settings, including hyperparameters.
\subsection{Evaluation}
\label{sec:evaluation}
As with ~\citet{feng_codebert_2020}, we use Mean Reciprocal Rank (MRR) as an evaluation metric.
\begin{equation*}
    \text{MRR} = \frac{1}{|Q|} \sum_{i=1}^{|Q|} \frac{1}{\text{rank}_i} \\
\end{equation*}
$|Q|$ refers to the total number of queries.
When a test set has 1,000 data pairs, given a natural language query$_i$, the model calculates the similarity with the corresponding code$_i$ and the 999 distractor codes.
If the similarity score given for code$_i$ is the 2nd highest among 1,000 codes, rank$_i$ equals 2.
Then, the average of the inverse of rank$_i$ over all queries and codes is calculated as MRR.

Table~\ref{table:dataset_size_csn} shows the sizes of CSN we use in our experiments.
Each test set of CSN for MRR evaluation contains 1,000 data pairs randomly sampled from the original test sets.
We use CoSQA and WQT as test sets in addition to CSN.
As well as CSN, we create CoSQA test sets from the original 20,604 data pairs. 
We compute the average of MRR scores over three different test sets for CSN and CoSQA.
The original WQT test set has 422 data pairs, so we use it as-is without sampling data like CoSQA.

We translate natural language queries in these test sets using the same machine translation model and parameter settings as the translation of the training data.
\subsection{Model Settings}
\label{sec:baseline_settings}
We prepare three model settings that differ in the amount and pattern of training data.
\paragraph{No-pre-training}
An XLM-R model with no further training applied and its initial parameters used.
\paragraph{All-to-One}
A model that uses data pairs of multilingual queries and monolingual codes for pre-training.
The size of pre-training data ranges from 1.2 million to 2.7 million, depending on programming languages.
\paragraph{All-to-All}
A model that uses data pairs of multilingual queries and multilingual codes for pre-training. 
The size of pre-training data is over 7.6 million.
\subsection{Results}
\label{sec:baseline_results}
Table~\ref{table:mrr_baseline} shows the scores of the MRR evaluation under all settings.
The scores with CSN showed that All-to-All performed best in Go, Java, and PHP in almost all natural languages.
On the other hand, All-to-One showed better scores than All-to-All on the Python test set.
It is possible that the performance reached the top at All-to-One on the Python test set, given that the difference in scores between All-to-One and All-to-All was relatively small (<0.1).
On CoSQA and WQT, there were also cases where model settings other than All-to-All performed better.

The performance of the original CodeBERT on a code search task is shown in Table~\ref{table:codebert_baseline}.
Overall, All-to-All is on par with the performance of CodeBERT in English data.
Especially, All-to-All marks better scores in Java and PHP than CodeBERT.
Note that our experiments and those of CodeBERT differ in the number of test sets used. 
Thus, it is difficult to compare these scores directly to discuss the model's superiority. 

We observed a gradual trend that the scores decreased in English and French and increased in Japanese and Chinese as we increased the size of the pre-training data.
This phenomenon might be due to the difference in knowledge of these languages acquired during pre-training XLM-R.
The XLM-R pre-training data contain approximately 350 GiB for English and French and approximately 69 GiB and 46 GiB for Japanese and Chinese, respectively.
As parameters of XLM-R were updated during our pre-training, the knowledge of English and French the model originally had was lost.
On the other hand, the scores of Japanese and Chinese, in which the model owned a small amount of data, were improved by increasing the data size.

\section{Analysis on Translation Quality}
\label{sec:analysis}
\subsection{Back-translation Filtering}
\label{sec:back-translation}
\begin{table}[t]
    \small\centering
    \tabcolsep = 0.06cm
    \begin{tabular}{ccccccc}
    \toprule
    & \multicolumn{6}{c}{Train} \\ 
    \cmidrule{2-7}
    & 0.2 & 0.3 & 0.4 & 0.5 & 0.6 & 0.7  \\
    \midrule
    FR&621,167&613,893&597,092&570,891&530,485&391,897\\ 
    JA&612,422&594,477&552,979&480,567&388,189&250,028\\ 
    ZH&607,468&588,808&557,748&500,622&410,369&265,986\\ 
    \midrule
    \midrule
    & \multicolumn{6}{c}{Valid} \\ 
    \cmidrule{2-7}
    & 0.2 & 0.3 & 0.4 & 0.5 & 0.6 & 0.7 \\
    \midrule 
    FR&27,881&27,535&26,799&25,621&24,000&20,231\\ 
    JA&27,433&26,524&24,901&21,981&16,327&10,304\\    
    ZH&27,115&26,178&24,971&22,280&18,445&10,792\\ 
    \bottomrule
    \end{tabular}
    \caption{The sizes of our dataset for fine-tuning after back-translation filtering applied.}
    \label{table:dataset_size_bt}
\end{table}
The translation quality of our dataset must affect the model's task performance.
Therefore, we investigate whether there is a difference in the scores of the code search task when we filter out the low-quality data from the fine-tuning dataset.

We apply a back-translation filtering method based on previous studies that used machine translation to automatically build a high-quality multilingual dataset from the English one ~\cite{sobrevilla-cabezudo-etal-2019-back,dou-etal-2020-dynamic,yoshikoshi_multilingualization_2020}.
We first apply back-translation to French, Japanese, and Chinese docstrings.
Then we calculate the uni-gram BLEU~\cite{papineni_bleu_2002} score between the back-translated docstrings and the original English ones and collect only data with scores higher than certain thresholds.
In our experiments, we conduct filtering to the fine-tuning dataset of Go.
Table~\ref{table:dataset_size_bt} shows the data sizes after back-translation filtering.
We set thresholds to 0.2 to 0.7 in increments of 0.1 and compare the model's performance with each threshold.
We choose these values because the sizes of the datasets change relatively hugely when filtered with the threshold 0.3 to 0.6 (Appendix~\ref{appendix:backtranslatoin_filtering}).
\begin{table}[t]
    \small\center
    \tabcolsep = 0.2cm
    \begin{tabular}{lccccccc}
    \toprule
    & 0 & 0.2 & 0.3 & 0.4 & 0.5 & 0.6 & 0.7 \\
    \midrule
    EN & .835 &  N/A & N/A & N/A &  N/A &  N/A &  N/A \\ 
    FR & .808 & .810 & .808 & .805 & \textbf{.811} & .809 & .807 \\ 
    JA & .816 & .805 & .803 & \textbf{.817} & .813 & .813 & .802 \\ 
    ZH & .804 & \textbf{.818} & \textbf{.818} & .807 & .798 & .802 & .802 \\ 
    \bottomrule
    \end{tabular}
      \caption{MRR scores with back translation filtering for fine-tuning data. 0 means no filtering applied.}
    \label{table:mrr_bt}
\end{table}
\subsection{Results}
\label{sec:bt_results}
Table~\ref{table:mrr_bt} shows the MRR scores of the models whose fine-tuning data are filtered with different thresholds.
In every language, the scores peak when we set the threshold between 0.2 to 0.5 and then drop with larger thresholds up to 0.7.
This result implies that the filtering successfully removes the low-quality data while maintaining the number of training data and leads to better MRR scores.
We assume that the change in size from the original dataset becomes more prominent with thresholds from 0.5 to 0.7  (around 100K-400K), thus eventually resulting in lowering the overall scores.

However, the score changes seem insignificant ($\pm{0.02}$) among these thresholds.
One possible reason is that the data size remains over 250K even after filtering, which should already be enough for fine-tuning in general.

In summary, the results show that filtering out some low-quality data improves the model's performance on the code search task, but removing over 150K data worsens the test scores. 

\section{Conclusion}
\label{sec:conclusion}
We created a large multilingual code search dataset by a neural machine translation model.
We then constructed a multilingual code search model using our dataset. 
We found out that the models pre-trained with all of the multilingual natural language and programming language data achieved the best performance on a code search task almost all the time.
We also investigated the relationship between the translation quality of our dataset and the model's performance.
The results indicated that the data size contributed more to the model's code search performance than the data translation quality.

Overall, this research introduced that using a publicly available machine translation model helps to translate texts in the programming domain.
We can apply our method to extend datasets for languages other than French, Japanese, and Chinese to construct models for various natural languages.


\section*{Limitations}
\label{sec:limitations}
We used XLM-R for the baseline model to train with our dataset in our experiments because we wanted to make experimental settings as close as the previous study of CodeBERT but for multilingual data.
Since CodeBERT is based on RoBERTa, we chose XLM-R, which is also RoBERTa-based and already trained with multilingual data.

\section*{Acknowledgements}
We thank the two anonymous reviewers for their helpful comments and suggestions, which improved this paper.
This research is supported by JSPS KAKENHI Grant Number JP20K19868 and partially by Microsoft Research Asia (Collaborative Research Sponsorship).


\bibliography{anthology,custom}
\bibliographystyle{acl_natbib}

\appendix
\begin{table*}[t]
    \scriptsize\centering
    \begin{tabular}{ccc}
    \toprule
        Original (EN) & Translated (JA) & Quality \\
    \midrule\midrule
    \begin{tabular}{l}
        SetStatus sets the Status field s value .
    \end{tabular}
    & 
    \begin{tabular}{l}
        \Ja{SetStatus は、Status フィールドの値を設定します。}
    \end{tabular}
    &  
        \cmark\\
    \midrule
    \begin{tabular}{l}
        retrieveCoinSupply fetches the coin supply \\
        data from the vins table .
    \end{tabular}
    & 
    \begin{tabular}{l}
        \Ja{retrieveCoinSupply は、vins テーブルから}\\
        \Ja{コイン供給データを取得します。}
    \end{tabular}
    &  
        \cmark\\
    \midrule
     \begin{tabular}{l}
        stateIdent scans an \underline{alphanumeric} or field .
    \end{tabular}
    & 
    \begin{tabular}{l}
        \Ja{stateIdent は、\underline{アルファナウマリ}または}\\
        \Ja{フィールドをスキャンします。}
    \end{tabular}
    &  
    \begin{tabular}{c}
        \xmark \\
        Unknown word
    \end{tabular}\\
    \midrule
     \begin{tabular}{l}
        VisitFrom calls the do function starting\\
        from the first neighbor \underline{w for which w $\ge$ a}\\ 
        \underline{with c equal to the cost of the edge}\\
        \underline{from v to w .} The neighbors are then\\
        visited in increasing numerical order .\\
        \underline{If do returns true VisitFrom returns}\\
        \underline{immediately skipping any remaining}\\
        \underline{neighbors and returns true .}
    \end{tabular}
    & 
    \begin{tabular}{l}
        \Ja{VisitFrom は、最初の隣人 w から始まる do 関数を}\\
        \Ja{呼び出し、\underline{その w $\ge$ a と c は v から w までの}}\\
        \Ja{\underline{エッジのコストに等しい。}}\\
        \Ja{\underline{If do returns true VisitFrom returns immediately}}\\
        \Ja{\underline{skipping any remaining neighbors and returns true.}}\\
        \Ja{もしそうであれば、VisitFromは直ちに}\\
        \Ja{残りの隣人を無視して true を返します。}
    \end{tabular}
    &  
    \begin{tabular}{c}
        \xmark \\
        Wrong translation / Omission
    \end{tabular}\\
    \bottomrule
    \end{tabular}
    \caption{Examples of query data from the dataset (Japanese, Go, threshold=0.4). These data are sampled from the top 10 entries of the dataset.}
    \label{table:appendix_data_examples}
\end{table*}

\begin{table*}[t]
    \scriptsize\centering
    \begin{tabular}{ccc}
    \toprule
    Original (EN) & Translated (JA) & Back-translated (EN) \\
    \midrule\midrule
    \begin{tabular}{l}
    NoError asserts that a function returned\\
    no error ( i . e . nil ) . \\
    \textbf{actualObj err : = SomeFunction ()}\\
    \textbf{if a . NoError ( err ) \{ assert .}\\
    \textbf{Equal ( t actualObj expectedObj ) \} }\\
    Returns whether the assertion \\
    was successful ( true ) or not ( false ) .
    \end{tabular}
    & 
    \begin{tabular}{l}
    \Ja{NoError は、関数がエラーを返しません}\\
    \Ja{( i. e. nil ) を主張します。}\\
    \underline{\Ja{まあ、あれ? まあ、あれ? まあ、あれ?}}\\
    \underline{\Ja{まあ、あれ? まあ、あれ? まあ、あれ?}}\\
    \Ja{真実(真実)か否かを返す。}
    \end{tabular}
    &  
    \begin{tabular}{l}
    NoError claims that the function \\
    does not return an error (i.e. nil).\\
    \underline{Oh well that? Oh well that? Oh well that?}\\
    \underline{Oh well that? Oh well that?}\\
    It is the truth or the truth.
    \end{tabular}\\
    \midrule
    \multicolumn{3}{c}{The original query contains a code-like sequence (bold texts), so the model could not successfully translate it (underline texts).} \\
    \bottomrule
    \end{tabular}
    \caption{An example of filtered-out query data (Japanese, Go, threshold=0.4).}
    \label{table:appendix_filtered_data}
\end{table*}

\begin{table*}[t]
    \centering
    \begin{tabular}{cccccccccc}
    \toprule
    & \multicolumn{9}{c}{Train} \\ 
    \cmidrule{2-10}
    & 0.1 & 0.2 & 0.3 & 0.4 & 0.5 & 0.6 & 0.7 & 0.8 & 0.9\\
    \midrule
    FR&626,130&621,167&613,893&597,092&570,891&530,485&391,897&224,928&78,989\\ 
    JA&621,857&612,422&594,477&552,979&480,567&388,189&250,028&76,965&27,670\\ 
    ZH&618,904&607,468&588,808&557,748&500,622&410,369&265,986&71,625&20,173\\ 
    \midrule
    \midrule
    & \multicolumn{9}{c}{Valid} \\ 
    \cmidrule{2-10}
    & 0.1 & 0.2 & 0.3 & 0.4 & 0.5 & 0.6 & 0.7 & 0.8 & 0.9\\
    \midrule 
    FR&28,123&27,881&27,535&26,799&25,621&24,000&20,231&11,646& 4,647\\ 
    JA&27,837&27,433&26,524&24,901&21,981&16,327&10,304&5,422& 1,806\\    
    ZH&27,693&27,115&26,178&24,971&22,280&18,445&10,792&4228&1,002\\ 
    \bottomrule
    \end{tabular}
    \caption{The sizes of our fine-tuning dataset after back-translation filtering with thresholds in increment of 0.1.}
    \label{table:appendix_dataset_size_bt}
\end{table*}

\section{CodeSearchNet}
\label{appendix:codesearchnet}
Table~\ref{table:dataset_size_codebert} shows the size of CSN for each programming language used for pre-training CodeBERT with MLM and fine-tuning on the code search task.
The number of data for fine-tuning in Go is listed as 635,635 in \citet{feng_codebert_2020}, but the dataset publicly provided contains 635,652 data.
\section{Dataset Translation}
\label{appendix:dataset_translation}
We manually evaluate the translation quality of our dataset.
Table~\ref{table:appendix_data_examples} shows examples of translation of query data from English to Japanese using M2M-100.
Since queries of CSN are based on source code descriptions, some of them contain strings that do not necessarily need to be translated, such as variable names, function names, and technical terms (e.g., \texttt{SetStatus}, \texttt{retrieveCoinSupply}).
M2M-100 successfully translates the entire sentence, leaving such domain-specific strings as needed.

On the other hand, we observe some errors, such as translating to unknown words (e.g., ``alphanumeric'' to \Ja{``アルファナウマリ''}) or omitting some texts from the translation.

We also manually annotate the labels of 45 sampled data pairs from the fine-tuning dataset of Japanese queries and Go codes and calculate how much they match the original labels.
These 45 data pairs do not contain queries that were not successfully translated and remain in English.
Among 45 data pairs, 28 of them have ``1'' as their labels and 17 for ``0''.
We calculate the correlation with accuracy, and the score is 0.911.
\section{Training Settings}
\label{appendix:training_settings}
As hyperparameters for pre-training the model, we set the batch size to 64, the maximum input length to 256, and the learning rate to 2e-4.
As hyperparameters for the fine-tuning of the model, we set the batch size to 16, the learning rate to 1e-5, and the number of max training epochs to 3.
In both cases, we use Adam as the optimizer.
\section{Back-translation Filtering}
\label{appendix:backtranslatoin_filtering}

Table~\ref{table:appendix_filtered_data} shows an example of the removed data by filtering. 
Table~\ref{table:appendix_dataset_size_bt} shows the data size of each filtering threshold. 

\end{document}